\pdfoutput=1

\documentclass[11pt]{article}
\usepackage[table]{xcolor}

\usepackage[]{ACL2023}

\usepackage{times}
\usepackage{latexsym}

\usepackage[T1]{fontenc}

\usepackage[utf8]{inputenc}

\usepackage{microtype}

\newcounter{notecounter}

\newcommand{\enoteson}{\long\gdef\enote##1##2{{
\stepcounter{notecounter}
{\large\bf \hspace{1cm}\arabic{notecounter} $<<<$ ##1: ##2 $>>>$\hspace{1cm}}}}}
\enoteson

\def\eqref#1{Eq.~\ref{eqn:#1}}

\usepackage{inconsolata}
\usepackage{tabularray}
\usepackage{graphicx}
\usepackage{hyperref}
\usepackage{amsmath}
\usepackage{bm}

\title{TED: Turn Emphasis with Dialogue Feature Attention \\
for Emotion Recognition in Conversation
}

\author{Junya Ono \and Hiromi Wakaki \\
       Sony Group Corporation \\
       \{junya.ono, hiromi.wakaki\}@sony.com}

\begin{document}
{\makeatletter\acl@finalcopytrue
  \maketitle
}
\begin{abstract}
Emotion recognition in conversation (ERC) has been attracting attention by methods for modeling multi-turn contexts. The multi-turn input to a pretraining model implicitly assumes that the current turn and other turns are distinguished during the training process by inserting special tokens into the input sequence. This paper proposes a priority-based attention method to distinguish each turn explicitly by adding dialogue features into the attention mechanism, called \textbf{T}urn
\textbf{E}mphasis with \textbf{D}ialogue (TED). It has a priority for each turn according to turn position and speaker information as dialogue features. It takes multi-head self-attention between turn-based vectors for multi-turn input and adjusts attention scores with the dialogue features. We evaluate TED on four typical benchmarks. The experimental results demonstrate that TED has high overall performance in all datasets and achieves state-of-the-art performance on IEMOCAP with numerous turns.
\end{abstract}

\section{Introduction}
\label{sec:sec1}
Emotion recognition in conversation (ERC) has been discussed for over decade \cite{DBLP:conf/webi/YangLC07, devlin-etal-2019-bert}. In ERC, an emotion label is estimated for the current turn from multiple turns’ worth of utterances and speaker information. Emotion understanding such as ERC has the potential to be used in chatbots, medical situations, and call centers. 

The methods of ERC are often used on the basis of past and future contexts, external commonsense knowledge, and speaker information \cite{Majumder_Poria_Hazarika_Mihalcea_Gelbukh_Cambria_2019, ghosal-etal-2020-cosmic, zhu-etal-2021-topic, shen-etal-2021-directed}. Especially, studies that involve modeling multiple turns have grown in number since the appearance of ERC datasets that embody  different perspectives including multiple speakers and topics \cite{DBLP:journals/corr/abs-1710-03957, DBLP:journals/corr/abs-1708-04299, DBLP:journals/corr/abs-1810-02508}.

\begin{figure}[h]
    \centering
    \includegraphics[width=0.5\textwidth]{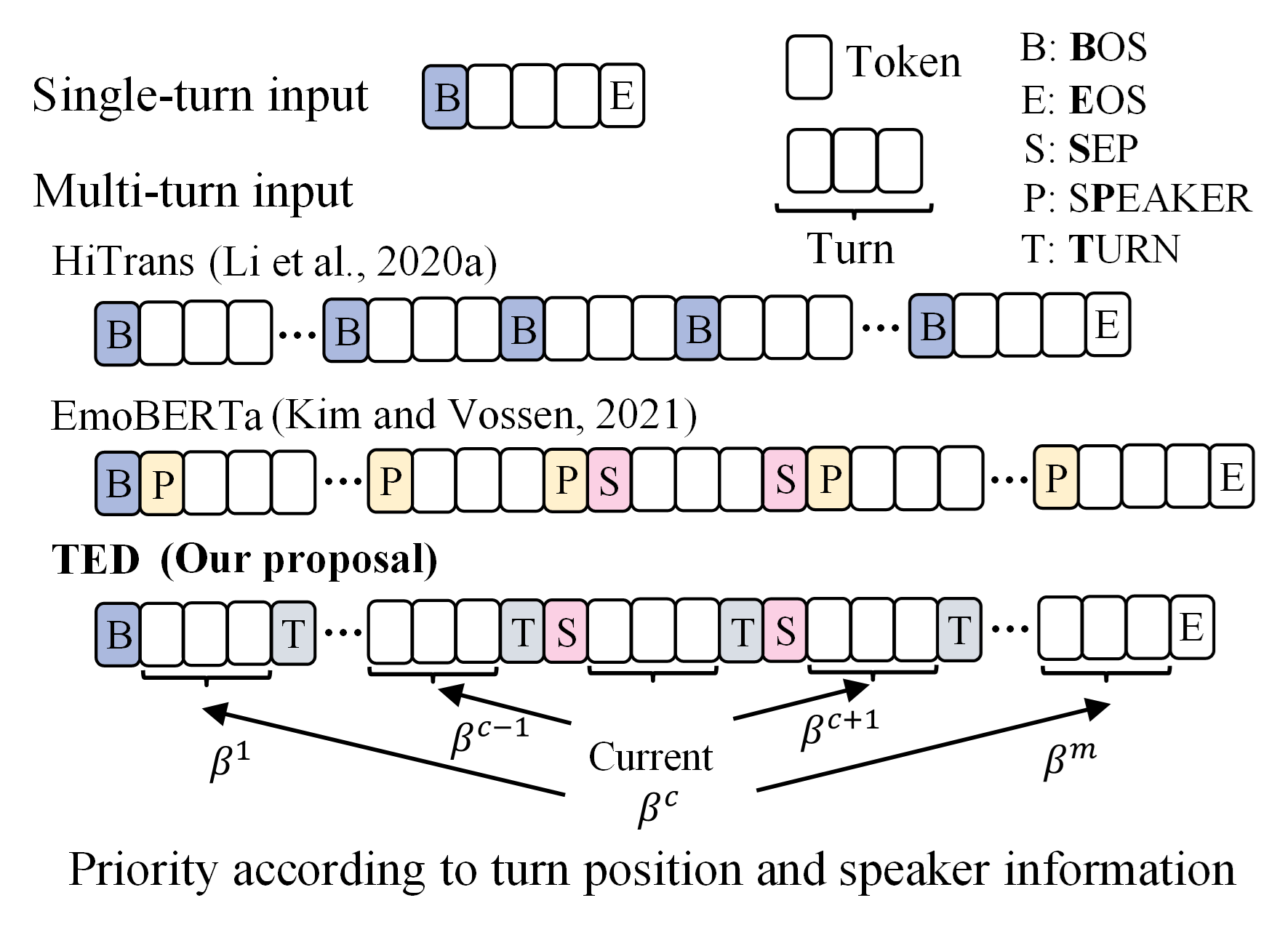}
    \caption{Concept of our method. This shows the difference in the multi-turn input in related methods. Our method prioritizes each turn to distinguish the turns}
    \label{fig:fig1}
\end{figure}

The mainstream multi-turn methods can exploit past and future turns for more contexts through their use of recurrent neural networks (RNNs), graph neural networks (GNNs), and Transformer \cite{DBLP:journals/corr/VaswaniSPUJGKP17}.
These methods often input the token sequence for only the current turn into a pretrained model, such as BERT \cite{devlin-etal-2019-bert} and RoBERTa \cite{DBLP:journals/corr/abs-1907-11692}. They cannot use the context other than the current turn for single-turn input. Therefore, the methods of multi-turn input have been recently proposed by inserting special tokens into the input sequence as shown in Figure \ref{fig:fig1} \cite{li-etal-2020-hitrans, DBLP:journals/corr/abs-2004-03588, DBLP:journals/corr/abs-2109-04008, DBLP:journals/corr/abs-2108-12009}. These methods can obtain deeper contexts by adding the utterances in multiple turns into the input sequence. On the other hand, they design to distinguish each turn by including information of turn position and speaker as special tokens. However, this multi-turn input implicitly expects to be distinguished between the current turn and other turns to be modeled in the process of machine learning.

\begin{figure}[h!]
    \centering
    \includegraphics[width=0.5\textwidth]{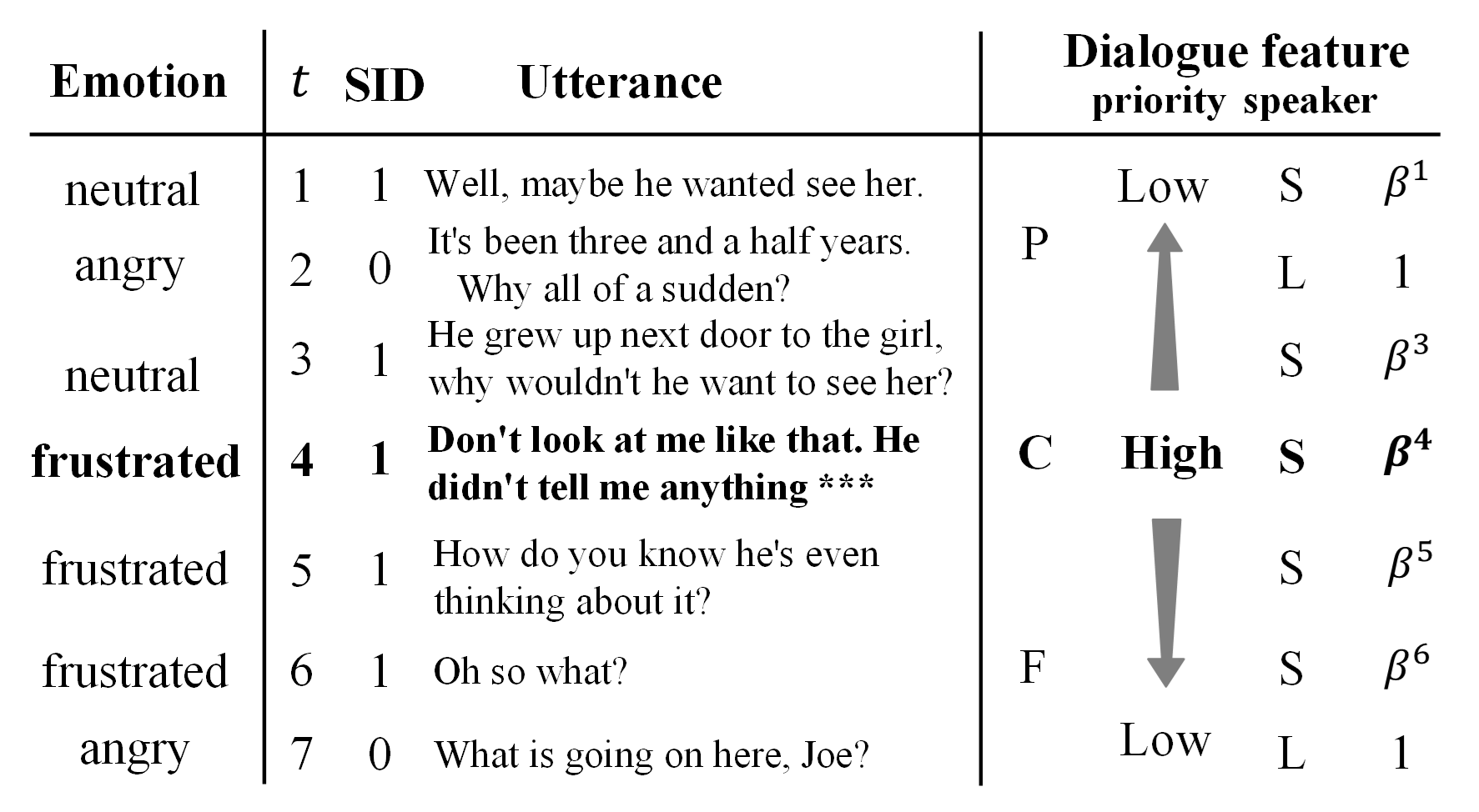}
    \caption{Example of ERC and dialogue features. TED uses current (C), past (P) and future (F) turns to obtain more contexts. Dialogue features indicate turn priority, same speaker (S) and listener (L). In this case, TED adjusts attention scores \(^{}\)for the same speakers (S) in the current turn by using an attention factor \(\beta^{t}\) with the turn priority; \({t}\) indicates the turn number; SID indicates the speaker identification.}
    \label{fig:fig2}
\end{figure}

We introduce a concept to distinguish explicitly between each turn and to control a degree of the distinction while making the most of dialogue features. As one of the methods, \ref{fig:fig1} shows our concept to clarify the distinction by prioritizing each turn and maximizing the priority of the current turn in comparison with related methods \cite{li-etal-2020-hitrans, DBLP:journals/corr/abs-2108-12009}. This concept uses special tokens of {[}TURN{]} and {[}SEP{]} as turn breaks and delimiter of the current turn, respectively. Moreover, it adjusts relationship between the current turn and other turns by using priority factor \(\beta^{t}\) calculated by dialogue features of turn position and speaker information; \({t}\) indicates position of a turn; SPK indicates Speaker information.

Our proposal, a priority-based attention method, called ``\textbf{T}urn \textbf{E}mphasis with \textbf{D}ialogue (TED)'', uses the above dialogue features.
First, TED creates a multi-turn input sequence by concatenating all turns with special tokens and also creates a turn-based vector by averaging the token-based vectors at the same positions of the utterance in a turn, called ``\textbf{T}urn-\textbf{B}ased \textbf{E}ncoding (TBE)''.
Second, it establishes multi-head self-attention (MHSA) between the turn-based vectors of TBE, called ``\textbf{T}urn-\textbf{B}ased \textbf{M}HSA (TBM)''.
Finally, it adds a dialogue layer to TBM to adjust attention scores for three types of all turns, the same speaker turns at the same as a current turn, or the listener turns. Figure \ref{fig:fig2} shows an example of making adjustments with an attention factor \(\beta^{t}\); \({t}\) indicates the turn number. The comprehensive results in four typical benchmarks demonstrate that TED has high performance in all datasets and especially archives state-of-the-art performance in IEMOCAP with a lot of turns.

\section{Related work}
\label{sec:sec2}
ERC has recently attracted attention because it can handle more complex contexts using multiple turns. Numerous powerful approaches have been proposed on the basis of RNNs, GNNs, and Transformers. Many use a pretrained model, such as BERT and RoBERTa, to make sequence representations corresponding to input tokens.

\subsection{Multi-turn models}
\label{sec:sec2-1}
Here, we review recent models based on three different neural networks.

\paragraph{RNN-based models} \ ICON \cite{hazarika-etal-2018-icon} and CMN \cite{hazarika-etal-2018-conversational} have gated recurrent unit (GRU) and memory networks. HiGRU \cite{jiao-etal-2019-higru} has two GRUs for the utterance and conversation. BiF-AGRU \cite{Jiao_Lyu_King_2020} has a hierarchical memory network with an attention GRU for historical utterances. DialogueRNN \cite{Majumder_Poria_Hazarika_Mihalcea_Gelbukh_Cambria_2019} utilizes the global state as a context and the party state for individual speakers by incorporating bidirectional GRUs for emotional dynamics. COSMIC \cite{ghosal-etal-2020-cosmic} has a similar structure to DialogueRNN but with added external commonsense knowledge. BiERU \cite{DBLP:journals/corr/abs-2006-00492} devised an efficient and party-ignorant framework by using a bi-recurrent unit. CESTa \cite{wang-etal-2020-contextualized} handles the global context by using Transformer and individual speakers by using BiLSTM-CRF. DialogueCRN \cite{DBLP:journals/corr/abs-2106-01978} has contextual reasoning networks that have long short-term memory (LSTM) to understand situations and speaker context.

\paragraph{GNN-based models} \ DialogGCN \cite{ghosal-etal-2019-dialoguegcn} handles the dependency and positional relationship of speakers as a graph structure. RGAT \cite{ishiwatari-etal-2020-relation} has a similar strategy to DialogGCN but with added positional encodings. ConGCN \cite{ijcai2019p752} builds an entire dataset including utterances and speakers as a large graph. SumAggGIN \cite{sheng-etal-2020-summarize} has two stages of summarization and aggregation graphs for capturing emotional fluctuation. DAG-ERC \cite{shen-etal-2021-directed} models the flow between long distance and nearby contexts. TUCORE-GCN \cite{DBLP:journals/corr/abs-2109-04008} constructs a graph of relational information in a dialogue with four types of nodes and three types of edges.

\begin{figure}[h!]
    \centering
    \includegraphics[width=0.5\textwidth]{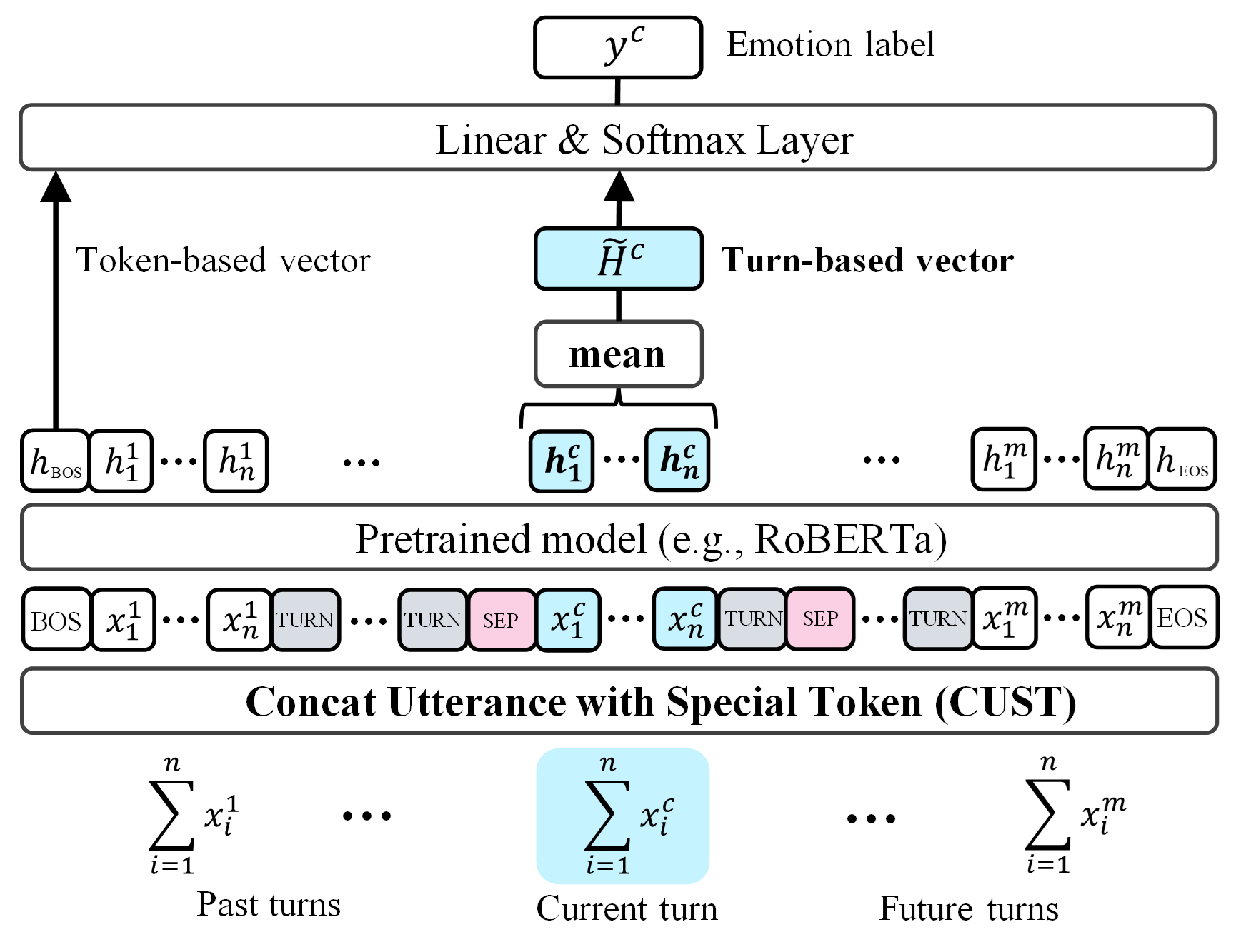}
    \caption{Turn-based encoding (TBE) model. CUST outputs a multi-turn sequence from utterances in past and future turns with ``TURN'' and ``SEP'' tokens as separators. TBE uses a current turn-based
vector\(\ {\widetilde{H}}^{c}\) created by averaging token-based vectors.}
    \label{fig:fig3}
\end{figure}

\paragraph{Transformer-based models} \ KET \cite{zhong-etal-2019-knowledge} incorporates hierarchical self-attention with commonsense knowledge. DialogueTRM \cite{DBLP:journals/corr/abs-2010-07637} uses a hierarchical Transformer and multi-grained network for multi-modal fusion. HiTrans \cite{li-etal-2020-hitrans} uses two hierarchical transformers, one for low-level representations from BERT, the other a high level turn-by-turn Transformer structure. DialogXL \cite{DBLP:journals/corr/abs-2012-08695} deals with long-term memory and four types of dialog-aware self-attention based on XLNet \cite{DBLP:journals/corr/abs-1906-08237}. TODKAT \cite{zhu-etal-2021-topic} integrates topic representations and commonsense knowledge into an encoder-decoder structure. EmoBERTa \cite{DBLP:journals/corr/abs-2108-12009} inputs one sentence including past and future turns with special tokens into RoBERTa.

\subsection{Input Encoding}
\label{sec:sec2-2}
Various formats have recently been used for expressing the multi-turn input by using utterances and speaker information in turns. To encode the multi-turn input sequence, one method adds speaker information in the form of a special token and adds segment, positional, and speaker embeddings to the utterance embedding \cite{DBLP:journals/corr/abs-2004-03588, DBLP:journals/corr/abs-1710-03957}. Another method replaces the speaker information with a real name and adds separator tokens to the front and back of the current turn \cite{DBLP:journals/corr/abs-2108-12009}. EmoBERTa, which is an example of the latter method, achieves high accuracy with only the token-based vector of the head token <s> and RoBERTa without a special multi-turn network. We believe that a richer context can be obtained from multi-turn utterances.

Our model, TED, is inspired by these methods, but it does not include speaker information in the input sequence. Instead, we apply speaker information to the attention mechanism.

\section{Methodology}
\label{sec:sec3}
The following three subsections describe TED as a classification problem using a pretrained model and multi-turn contexts.

\subsection{Turn-Based Encoding}
\label{sec:sec3-1}
Here, we introduce the turn-based encoding model (TBE), which has two components, i.e.\, input encoding and creation of the mean vector of utterance to encode multi-turn utterances. Figure \ref{fig:fig3} shows the structure of TBE.

\paragraph{Input Encoding} First, TBE creates token-based vectors by concatenating multi-turn utterances including past and future turns with special tokens (called “Concat Utterance with Special Token”, or CUST). 
As indicated in Figure 3, CUST outputs a token sequence including past and future turns as follows:
\begin{equation}
    CUST\ : = Concat\left( {\bm{x}^{c}\bm{,X}}^{p},\ \bm{X}^{f},t_{T},t_{s} \right)
    \label{eq:eq1}
\end{equation}
where x denotes a token in an utterance; \(\bm{x}^{c} = \sum_{i}^{}x_{i}^{c}\) denotes the token sequence of an utterance at the current turn; \(\bm{X}^{p} = \left(\sum_{i}^{}x_{i}^{1} \text{ ,\(\cdots\), } \sum_{i}^{}{x_{i}^{c - 1}}\right) \) and \(\bm{X}^{f} = \left(\sum_{i}^{}{x_{i}^{c + 1} \text{ ,\(\cdots\), }} \sum_{i}^{}{x_{i}^{m}}\right)\) denote the list of the token sequences in past and future turns, respectively; \({t}_T\) and \({t}_S\) denote special tokens [TURN] and [SEP], respectively. CUST outputs a token sequence by adding [TURN] at the end of each turn to distinguish turn breaks and by adding [SEP] to the front and back of the current turn to distinguish it from the other turns. [SEP] is also added to the end of a sequence when pretraining. By comparison, HiTrans uses a CLS token at the beginning of each turn to distinguish between turns. However, the CLS token is often used as a context vector in a downstream task as the first token of BERT. Therefore, we add a new token [TURN], which means only turn breaks.

\begin{figure}[h!]
    \centering
    \includegraphics[width=0.5\textwidth]{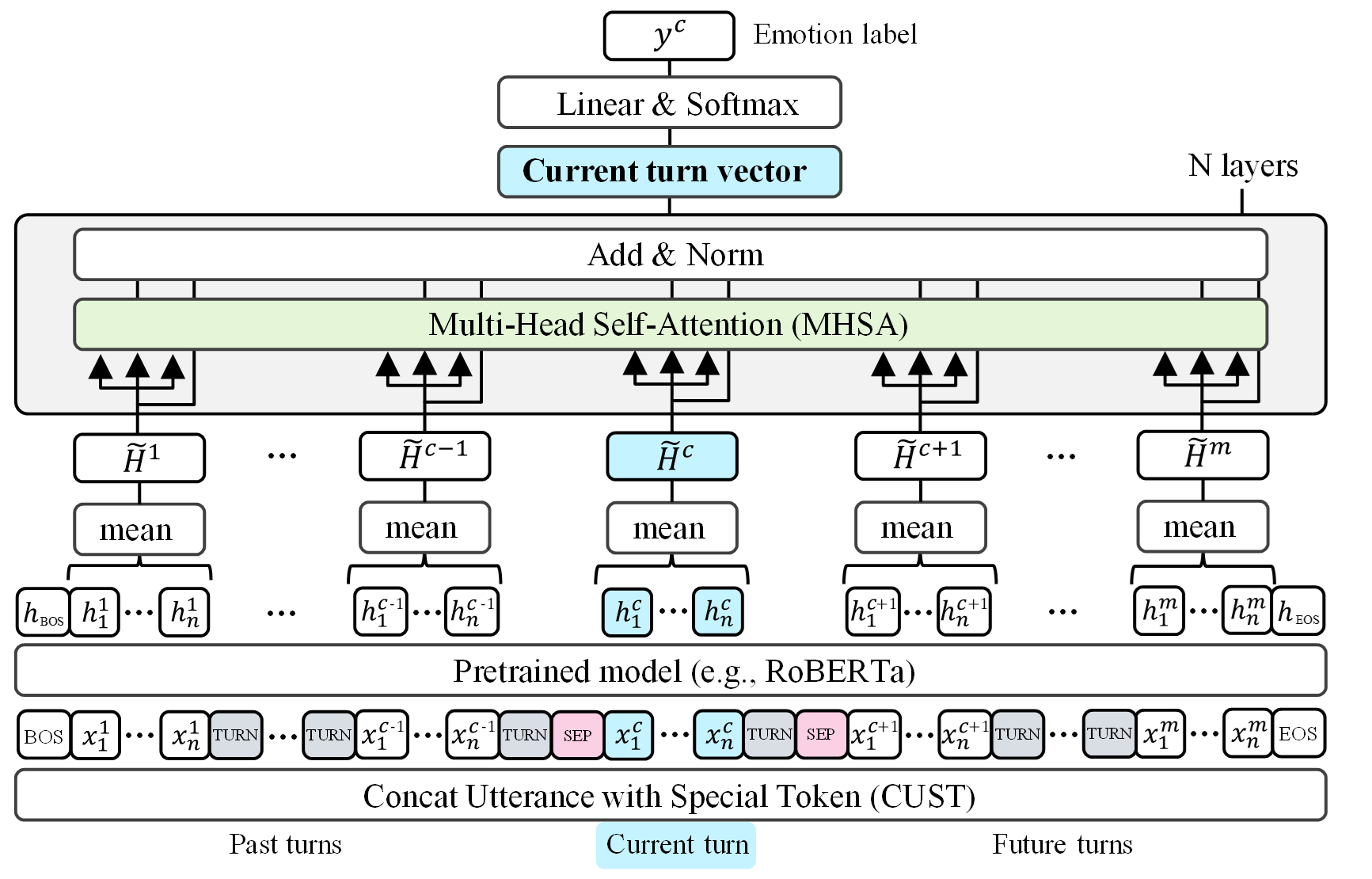}
    \caption{Turn-based MHSA (TBM) model. TBM establishes MHSA between turn-based vectors to obtain more contexts based on TBE.}
    \label{fig:fig4}
\end{figure}

The “pretrained model” (PTM) in Figure \ref{fig:fig3} is a pretrained model such as RoBERTa that outputs a list \(\mathbf{H}\) of token-based vectors per turn for all turns as follows:

\begin{equation}
\begin{split}
    \bm{H} \ \mathbf{=} \ \left( \bm{h}_{\ \ ,\cdots,}^{1}\bm{h}_{\ \ ,\cdots,}^{c}\bm{h}^{m} \right) \hspace{2cm} \\
    where \ \bm{h}^{c} = \left( h_{1,\cdots,}^{c}h_{i,\cdots,}^{c}h_{n}^{c} \right)
    \label{eq:eq2}
\end{split}
\end{equation}
where \(\mathbf{h}^{c}\) denotes the list of token-based vectors in the current turn, where the range of tokens exactly matches the positions of the input sequence in the current turn among the outputs of PTM. Note that the vectors corresponding to the BOS and end EOS tokens are omitted.

\paragraph{Mean Vector of Utterance} \ Second, TBE creates turn-based vectors in order to average the token-based vectors corresponding to the utterance positions in each turn. 

The list of the turn-based vectors \(\widetilde{\bm{H}}\) in all m turns is defined as
\begin{equation}
\begin{split}
    \widetilde{\bm{H}} \ \mathbf{=} \ \left( {\widetilde{H}}_{\ ,\ldots,}^{1}{\widetilde{\ H}}_{\ ,\ldots,}^{c}{\widetilde{\ H}}_{\ }^{m} \right) \hspace{1cm} \\
    where \ {\widetilde{H}}^{c} = Mean(\bm{h}^{c})
    \label{eq:eq3}
\end{split}
\end{equation}
where \(\widetilde{H}^{c}\) in Figure \ref{fig:fig3} is the average of the \(\bm{h}^{c}\). The label with the highest probability \(\bm{P}\) among \(\bm{L}\) labels is chosen as the “Emotion label” in Figure \ref{fig:fig3}:
\begin{equation}
\begin{split}
    \bm{P} & = \left( P_{l,\cdots,}P_{l,\cdots,}P_{L} \right) \\
    & = Softmax \left\{ F\left( \widetilde{H}^{c}\right) \right\} \\
    where & \ P_l = P\left( y_l|\bm{x}^c, \bm{X}^p, \bm{X}^f, s^c, \bm{S}^p, \bm{S}^f \right) \hspace{-1cm}
    \label{eq:eq4}
\end{split}
\end{equation}
where \(F\) is a linear function; \(s^{c}\) denotes speaker information at the current turn;
\(\bm{S}^{p} = {(s}_{\ }^{1}\text{ ,\(\cdots\), }s^{c - 1})\) and \(\bm{S}^{f} = {(s}_{\ }^{c + 1}\text{ ,\(\cdots\), }s^{m})\);denote the list of speaker information of past and future turns;\(\ y_{l}\) denotes the \(l\)\textsuperscript{th} emotion label.

\begin{figure}[h!]
    \centering
    \includegraphics[width=0.5\textwidth]{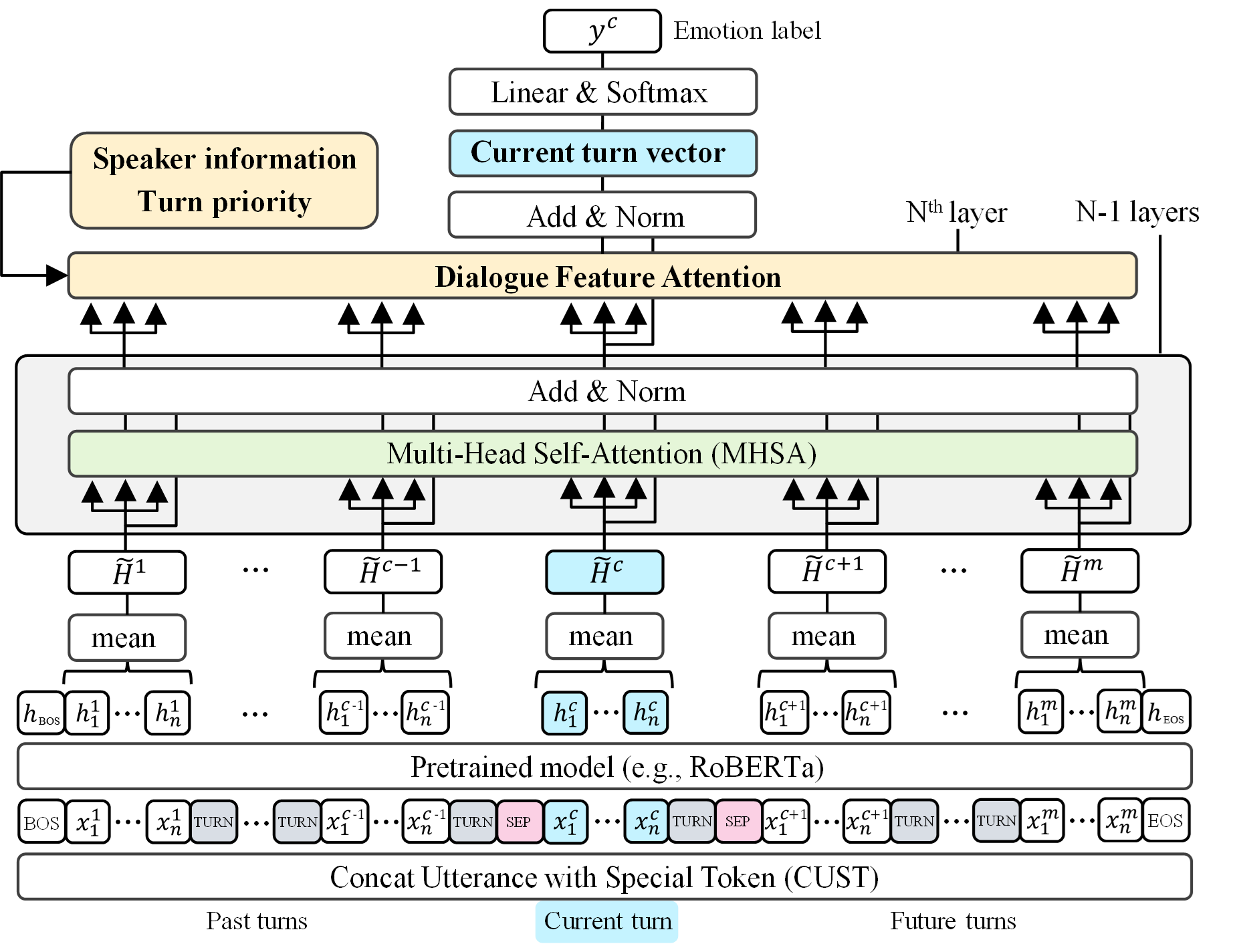}
    \caption{The proposed model, TED, has a dialogue layer to adjust attention scores by using turn priority and speaker IDs at the last (\({N}\)\textsuperscript{th}) layer on the basis of TBM.}
    \label{fig:fig5}
\end{figure}

\subsection{Turn-Based Multi-Head Self-Attention}
\label{sec:sec3-2}
We use the multi-head self-attention (MHSA) of Transformer \cite{DBLP:journals/corr/VaswaniSPUJGKP17} between turns to obtain more context based on TBE. Figure \ref{fig:fig4} shows our turn-based MHSA (TBM) model. The output of MHSA is added to the original input; then, the layer norm is taken in the same way as in the vanilla Transformer.

TUCORE-GCN uses MHSA masked with a turn-based window size and performs attention on a token-by-token basis. HiTrans also establishes token-based MHSA between CLS tokens, which is added to the input at the beginning of each turn. Our model differs from TUCORE-GCN in that it is MHSA with turn-based vectors of positions that exactly match the tokens of utterances.

\subsection{Turn Emphasis with Dialogue (TED)}
\label{sec:sec3-3}
We propose ``\textbf{T}urn \textbf{E}mphasis with \textbf{D}ialogue
(TED)'' to emphasis the current turn with two dialogue features, called ``priority`` and ``speaker``, as shown Figure \ref{fig:fig2}. We add a dialogue layer to TBM to adjust the attention scores with the turn priority and speaker information, as shown in Figure \ref{fig:fig5}.

The multi-head self-attention in Figure 5 has the same structure as the vanilla Transformer. \({\widetilde{\mathbf{\alpha}}}_{j}\) is the attention score matrix for all m turns at the \(j\)\textsuperscript{th} head of MHSA as follows:

\begin{equation}
\begin{split}
    \widetilde{\bf{\alpha}}_{j} \ \mathbf{=} \ \left( {\widetilde{\alpha}}_{j\ ,\ldots,}^{\ 1}{\widetilde{\ \alpha}}_{j\ ,\ldots,}^{\ t}{\widetilde{\ \alpha}}_{j\ }^{\ m} \right) \hspace{1cm} \\
    where \ {\widetilde{\alpha}}_{j}^{\ t} = \frac{exp({\alpha}_j^{\ t})}{\sum_{k}exp({\alpha}_j^{\ k})}
    \label{eq:eq5}
\end{split}
\end{equation}

where \({\widetilde{\alpha}}_{j}^{\ t}\) denotes the attention score vector for the \(t\)\textsuperscript{th} turn after the softmax function.

MHSA outputs attentional vectors by a concatenation with each head as follows:

\begin{equation}
\begin{split}
    MHSA := Concat & \left( {\bm{A}}_{1\ ,\ldots,}\ \bm{A}_{j\ ,\ldots,}^{\ t}\ \bm{A}_{J}^{\ m} \right) \\
    where \ \bm{A}_j = & \left( A_{j\ ,\ldots,}^{\ 1}\ A_{j\ ,\ldots,}^{\ t}\ A_{j}^{\ m}\right) \hspace{-4cm} \\
    A_j^{\ t} \ = & \ \widetilde{\alpha}_j^{\ t}\ \bm{V}_j\ , \ \bm{V}_{j} = \widetilde{\bm{H}} W_j^{V}
    \label{eq:eq6}
\end{split}
\end{equation}

where \(\widetilde{\bm{H}}\) denotes the list of turn-based vectors for all \({m}\) turns, \({W_j^{V}}\) denote learning parameters to scale the dimension of the model and the head, respectively, \(\bm{V_j}\) denotes the list of scaled vectors for all turns, and \(A_j^{\ t}\) denotes the attentional vector using the above attention scores.

So far, all turns are treated equally. Therefore, to emphasize the current turn, we construct a new dialogue layer, as follows. Let \(\widetilde{\bm{\alpha}}_{j}^{\ t\ '}\) be the weighted score vector produced by the dialogue feature attention in the last (\({N}\)\textsuperscript{th}) layer in Figure \ref{fig:fig5}.

\begin{equation}
    \widetilde{\bm{\alpha}}_{j}^{\ t\ '}\ \mathbf{=} \frac{\beta^t\ exp\left(\alpha_j^{\ t}\right)}{\sum_k\ \beta^{k}\ exp\left(\alpha_j^{\ k}\right)}
    \label{eq:eq7}
\end{equation}

where \({\beta}^t\) denotes the attention factor at the \({t}\)\textsuperscript{th} turn, which is calculated by the following two points.   

\paragraph{Turn priority} \  The current turn is emphasized with the priority shown in Figure \ref{fig:fig2}. The dialogue feature attention adjusts the priority in accordance with the following normal distribution.

\footnotesize
\begin{equation}
    \beta^t := \gamma^t = 
    \left\{
      \begin{split}
         & \gamma & {\tiny Constant} \\
         1 + \gamma \ exp & \left\{ - \frac{(t - t^c)^2}{2{\sigma}^2} \right\} & {\tiny Normdist}
      \end{split}
   \right.
   \label{eq:eq8}
\end{equation}
\normalsize

where \({\gamma}^t\) denotes either no priority (i.e., \({Constant}\)) or a variable priority based on a normal distribution (\({Normdist}\)), \(\gamma\) denotes the maximum coefficient, and \({t}^c\) and \(\sigma\) respectively denote the current turn and standard deviation. In case of \({Normdist}\), Formula (\ref{eq:eq8}) shows that the current turn is emphasized by the turn priority; the farther away the turn is from the current turn \(t^c\), the smaller the value of \({\gamma}^t\) (i.e., \({Normdist}\)) becomes.

  
\begin{table}[h]
\small
\centering
\begin{tblr}{
  width = \linewidth,
  colspec = {Q[1.2]Q[1]Q[1]Q[1.4]},
  columns = {halign=c, valign=m},
  hline{1-2,6} = {-}{},
}
\textbf{Dataset}  & \textbf{\# ALL dialogues} & \textbf{\# All turns} & \textbf{Std. turns in a dialogue} \\
IEMOCAP     & 151       & 7,433     & 16.8  \\
MELD        & 1,432     & 13,708    & 5.79  \\
EmoryNLP    & 827       & 9,489     & 5.34  \\
DailyDialog & 13,118    & 102,979   & 3.99   
\end{tblr}
\caption{Statistics of four ERC datasets}
\label{tab:tab1}
\end{table}

\paragraph{Speaker Information} \ speaker IDs are applied to the attention mechanism. The dialogue feature attention adjusts the attention factor for the same speaker or listener as follows:

\begin{equation}
    \beta^t =
    \left\{
    \begin{split}
       {\gamma}^t \ , \quad& \ S^t = S^c \\
       1 \ , \quad& otherwise
    \end{split}
    \right.
    \label{eq:eq9}
\end{equation}
\begin{equation}
    \beta^t =
    \left\{
    \begin{split}
       {\gamma}^t \ , \quad& \ S^t \neq S^c \\
       1 \ , \quad& otherwise
    \end{split}
    \right.
    \label{eq:eq10}
\end{equation}

where \(S^c\) and \(S^t\) denote the speaker IDs of the current and \({t}\)\textsuperscript{th} turn, respectively. Formula (\ref{eq:eq9}) is used to weight the turns that have the same speaker as the current turn and Formula (\ref{eq:eq10}) is used to weight the listener turns.

\section{Experiment setting}
\label{sec:sec4}
We evaluate TED on four ERC datasets. The datasets include speaker IDs for every turn. The training, development, and test data split follow the related work, such as COSMIC \cite{ghosal-etal-2020-cosmic}.

\subsection{Dataset}
\label{sec:sec4-1}
Table \ref{tab:tab1} shows the statistics of the datasets. IEMOCAP \cite{busso2008iemocap} is a multimodal dataset including text transcriptions of two speakers and it is annotated with six emotions. \\
\textbf{MELD} \cite{DBLP:journals/corr/abs-1810-02508} is a multimodal dataset
created from a TV show, Friends. It includes 260 speakers and is labeled with seven emotions. \\
\textbf{EmoryNLP} \cite{DBLP:journals/corr/abs-1708-04299} uses the same data source as MELD, i.e., the TV show Friends, and it annotates different utterances as compared with MELD by seven emotions. The utterances are from 225 speakers. \\
\textbf{DailyDialog} \cite{DBLP:journals/corr/abs-1710-03957} contains utterances of two speakers communicating on various topics related to daily life. It is annotated with seven emotions.

\begin{table*}[ht]
\small
\centering
\begin{tblr}{
  width = \linewidth,
  colspec = {Q[2]Q[1.2]Q[1.1]Q[1.2]Q[1.1]Q[1.2]Q[1.1]Q[1.2]Q[1.1]},
  columns = {halign=c, valign=m},
  cell{1}{1} = {r=2, c=1}{halign=c, valign=m},
  cell{1}{2} = {r=1, c=2}{halign=c, valign=m},
  cell{1}{4} = {r=1, c=2}{halign=c, valign=m},
  cell{1}{6} = {r=1, c=2}{halign=c, valign=m},
  cell{1}{8} = {r=1, c=2}{halign=c, valign=m},
  hline{1,3,11,12} = {-}{},
}
Model               & IEMOCAP   &           & MELD      &           & EmoryNLP  &           & DailyDialog   &           \\
                    & W-Avg F1  & Micro F1  & W-Avg F1  & Micro F1  & W-Avg F1  & Micro F1  & W-Avg F1      & Micro F1  \\
HiTrans             & 64.50     & -         & 61.94 & -         & 36.75     & -         & 54.93     & -  \\
DialogX             & 65.94     & -         & 62.41 & -         & 34.73     & -         & 54.93     & -  \\
COSMIC              & 65.28     & -         & 65.21 & -         & 38.11     & -         & 58.48     & 51.05  \\
CESTa               & 67.10     & -         & 58.36 & -         & -         & -         & \textbf{63.12}  & -      \\
DAG-ERC             & 68.03     & -         & 63.65 & -         & 39.02     & -         & 59.33     & -  \\
EmoBERTa            & 68.57     & -         & 65.61 & -         & -         & -         & -         & -      \\
TODKAT              & 61.33     & 62.60 & \textbf{68.23} & 64.75 & \textbf{43.12} & 42.68 & 58.47 & 52.56 \\
TUCORE-GCN          & -         & -         & 65.36 & -         & 39.24     & -         & 61.91     & -  \\

\textbf{TED (Ours)} & \textbf{68.63} & \textbf{68.82} & 66.35 & \textbf{67.25} & 39.07 & \textbf{43.33} & 62.43 & \textbf{55.71} &           \\
\end{tblr}
\caption{Overall comparison with state-of-the-art models. TED had good overall performance and achieved the state-of-the-art performance for IEMOCAP with a lot of turns.}
\label{tab:tab2}
\end{table*}

IEMOCAP is different from the other datasets in that its standard deviation (Std.) of the number of turns in one dialogue is 16.8, while the Std. of the others range from 3.99-5.79. This means it is possible to use more surrounding contexts for IEMOCAP. Regarding the speaker IDs of MELD and EmoryNLP, we assign new IDs to each dialogue to avoid the low frequency of appearance due to the large number of speakers

\subsection{Training and evaluation setting}
\label{sec:sec4-2}
We use RoBERTa-large of the Hugging Face library \cite{wolf-etal-2020-transformers}.

Each experiment consists of five trials and the average values of the trials are reported so that the initial values of parameters do not affect the judgments about the models. Moreover, each trial has different seeded fixed values and ran under the same computer specifications in order to maintain reproducibility.

As in the related work, we calculate micro average F1 (Micro F1) excluding the majority neutral and macro average F1 (Macro F1) on all labels for DailyDialog. On the other datasets, we calculate the weighted average f1 (W-Avg f1) and micro F1 for all labels. The evaluation scores on the test data are calculated for the model of the epoch with the highest score in the development data.

All experiments are subject to the same conditional schedule of the learning rate and early stopping. The learning rate is multiplied by a fixed value (0.8) when the evaluation score of the development data does not increase at the end of the epoch. Early stopping forcibly terminates the learning when the best score are not updated within 5 epochs

Regarding the model parameters, we use two attention layers and four heads. In Formula (\ref{eq:eq8}), \(\sigma\) is the standard deviation of the number of turns in all dialogues as shown in Table \ref{tab:tab1} and \(\gamma\) is a hyperparameter for the turn priority; it was set to 1.5, 2, 3, and 5.
 
\section{Results and Analysis}
\label{sec:sec5}
\subsection{Overall Performance}
\label{sec:sec5-1}

Table \ref{tab:tab2} compares the performance of TED and the other latest models listed in Section \ref{sec:sec2}. TED had good overall performance, while the other models had good performance on certain datasets. Moreover, TED had an advantage in terms of the dialogue features on datasets with a lot of turns, such as IEMOCAP.

In Formula (\ref{eq:eq7}), the attention score vector is a function of

\begin{itemize}
\item all turns, turns that have the same speaker as in the current turn, or listener turn.
\item a decay factor determined by turn priority (No decay (i.e., \({Constant}\)) or a Normal distribution centered on a current turn (i.e., \({Normdist}\)).
\end{itemize}

In terms of the context turn in CUST, we target the past only or both the past and the future. We describe the above combination of parameters of the results in Table \ref{tab:tab2}. The IEMOCAP column in Table \ref{tab:tab2} shows results for past, Listener, and Normdist; the MELD column those for past, Listener, and Constant; the EmoryNLP column those for both the past and the future, All turns, and Normdist, and the DailyDialog those for both the past and the future, Listener, and Constant.

The results suggest that TED strengthens the relationships between turns and accelerates the turn emphasis by specially treating the current turn.

\begin{table}[h!]
\footnotesize
\centering
\begin{tblr}{
  width = \linewidth,
  colspec = {Q[0.9]Q[1.5]Q[1.2]Q[1.2]Q[1.2]Q[0.8]},
  columns = {halign=c, valign=m},
  cell{1}{1} = {r=2, c=1}{halign=r, valign=m},
  cell{3}{1} = {r=6, c=1}{halign=c, valign=m},
  cell{9}{1} = {r=6, c=1}{halign=c, valign=m},
  cell{1}{2} = {r=2, c=1}{halign=r, valign=m},
  hline{1,3, 15} = {-}{},
  hline{9} = {2-6}{},
}
{\bf{Context}\\\bf{turn}} & {\bf{Dialog}\\\bf{feature}} & \textbf{IE} & \textbf{ME} & \textbf{EN} & \textbf{DD}  \\
                     &             & \textbf{W-Avg F1} & \textbf{W-Avg F1} & \textbf{W-Avg F1} &  \textbf{Micro F1}    \\
past    & TBM       & 68.50 & 65.87 & 38.09 & 61.52 \\
        & \ + A, N  & 68.56 & 66.05 & 38.60 & 61.09 \\
        & \ + S, C  & 68.48 & 65.94 & 38.78 & 61.20 \\
        & \ + S, N  & 68.59 & 66.25 & 38.56 & 61.10 \\
        & \ + L, C  & 68.12 & \textbf{66.35} & 38.64 & 61.08 \\
        & \ + L, N  & \textbf{68.63} & 66.10 & 38.59 & 61.18 \\
{past\\+\\future}   & TBM      & 67.62 & 65.60 & 38.93 & 61.99 \\
                    & \ + A, N & 68.13 & 65.96 & \textbf{39.07} & 62.16 \\
                    & \ + S, C & 67.77 & 66.11 & 38.96 & 62.39 \\
                    & \ + S, N & 68.02 & 65.89 & 38.64 &  62.20\\
                    & \ + L, C & 68.02 & 65.86 & 38.87 & \textbf{62.43} \\
                    & \ + L, N & 68.13 & 66.03 & 38.89 & 62.20 \\
\end{tblr}
\caption{TED performance for different combinations of dialogue items.  Dialog features (DF) are All turns (A), Same speaker turns (S), Listener turns (L), Constant (C), and Normdist (N) on IEMOCAP (IE), MELD (ME), EmoryNLP (EN, and DailyDialog (DD).}
\label{tab:tab3}
\end{table}

\begin{table}[h!]
\footnotesize
\centering
\begin{tblr}{
  width = \linewidth,
  colspec = {Q[2.3]Q[1.1]Q[1.4]Q[1.4]},
  cell{1}{1} = {r=2, c=1}{halign=c, valign=m},
  cell{1}{2} = {r=2, c=1}{halign=C, valign=m},
  cell{4}{1} = {r=3, c=1}{halign=c, valign=m},
  cell{7}{1} = {r=3, c=1}{halign=c, valign=m},
  columns = {halign=c, valign=m},
  hline{1,3,10} = {-}{},
  hline{4,7} = {2-4}{},
}
\bf{Model} & {\bf{Context}\\\bf{turn}} & \bf{IEMOCAP}  & \bf{MELD} \\
                &                       & \bf{W-Avg F1} & \bf{W-Avg F1} \\
Baseline        & current               & 55.76             & 64.27 \\
{TBE\\(Input Encoding)} & w/ past         & 66.20             & 65.11 \\
                      & w/ future       & 62.53             & 64.18 \\
                     & w/ past + future & 63.91            & 65.34 \\
{TBE\\(Input Encoding +\\Mean Vector of Utterance} & w/ past & \bf{66.70} & \bf{65.66} \\
                      & w/ future       & 64.17             & 64.75 \\
                      & w/ past + future & 66.62            & 65.52
\end{tblr}
\caption{Comparison of Baseline and TBE model. Baseline uses only the current turn and BOS token vector.}
\label{tab:tab4}
\end{table}

\begin{table}[h]
\footnotesize
\centering
\begin{tblr}{
  width = \linewidth,
  colspec = {Q[0.9]Q[2.7]Q[1.4]Q[1.4]},
  columns = {halign=c, valign=m},
  cell{1}{1} = {r=2, c=1}{halign=c, valign=m},
  cell{1}{2} = {r=2, c=1}{halign=C, valign=m},
  cell{3}{1} = {r=6, c=1}{halign=c, valign=m},
  cell{9}{1} = {r=6, c=1}{halign=c, valign=m},
  column{2} = {halign=l},
  hline{1,3,15} = {-}{},
  hline{9} = {2-4}{},
}
{\bf{Context}\\\bf{turn}} & \bf{\ Diaglog feature} & \bf{IEMOCAP}  & \bf{MELD} \\
                &                       & \bf{W-Avg F1} & \bf{W-Avg F1} \\
past & TED (Table \ref{tab:tab2})  & \bf{68.63}             & \bf{66.35} \\
     & TBE (Table \ref{tab:tab4})  & 66.70             & 65.66 \\
     & TBM  & 68.50            & 65.87 \\
     & \ + Speaker token  & 67.44            & 65.50 \\
     & \ + Speaker attention  & 68.40            & 65.51 \\
     & \ + Listener attention  & 68.40            & 65.85 \\
{past\\+\\future} & TED (Table \ref{tab:tab2})  & 68.13             & 66.11 \\
     & TBE (Table \ref{tab:tab4})  & 66.62             & 65.52 \\
     & TBM  & 67.62            & 65.69 \\
     & \ + Speaker token  &  66.67            & 65.25 \\
     & \ + Speaker attention  &  67.58            & 66.18 \\
     & \ + Listener attention  & 67.43            & 65.57
\end{tblr}
\caption{Comparison with related methods using speaker information. Neither adding token (Speaker token) nor limiting the attentional turns (Speaker and Listener attention) has an advantage.}
\label{tab:tab5}
\end{table}

\subsection{Detailed Performance}
\label{sec:sec5-2}
Here, we examine the results of TED in more detail by analyzing the effect of different parameters in the dialogue feature attention. In Table \ref{tab:tab3}, the ``Dialogue Feature`` column indicates the combination of two dialog feature items; A, S, and L indicate the target turns to be weighted by all turns (A), the same speaker turns (S), and listener turns (L), respectively; ``C`` indicates no decay factor (i.e., \({Constant}\)), while ``N`` indicates the normal distribution in Formula (\ref{eq:eq8}).

The results for the different context turns in Table \ref{tab:tab3} vary depending on the dataset. MELD and EmoryNLP have similar data statistics, as shown in Table \ref{tab:tab1}. Nevertheless, TED’s performance was higher in the context turn of the past only for MELD but was higher in the context turn of both the past and future for EmoryNLP.

The dialogue feature attention provided a slight improvement compared with that of TBM, which does not use the dialogue features. Varying the combinations of features yielded no significant differences in the results.

\subsection{Effect of Input Encoding and Mean Vector of Utterance (TBE)}
\label{sec:sec5-3}
Table \ref{tab:tab4} compares the TBE models that have only input encoding part or both the input encoding part and mean vector of utterance part with a baseline that does not have either part. The ``Context turn`` column indicates whether to include past, future, or both kinds of turn in CUST; ``no use`` indicates only the current turn for the baseline.

The results show the effectiveness of the past turns and TBE. The future turns do not contribute to a performance improvement. We consider that the current state is influenced by the past states in a chain reaction, while the current state is not directly affected by the future states.

The performance on IEMOCAP is significantly improved in the case of past turns. The performance of EmoBERTa shows the same tendency as TBE on IEMOCAP.

These results suggest that the turn-based vector of TBE is effective.

\begin{figure}[h!]
    \centering
    \includegraphics[width=0.5\textwidth]{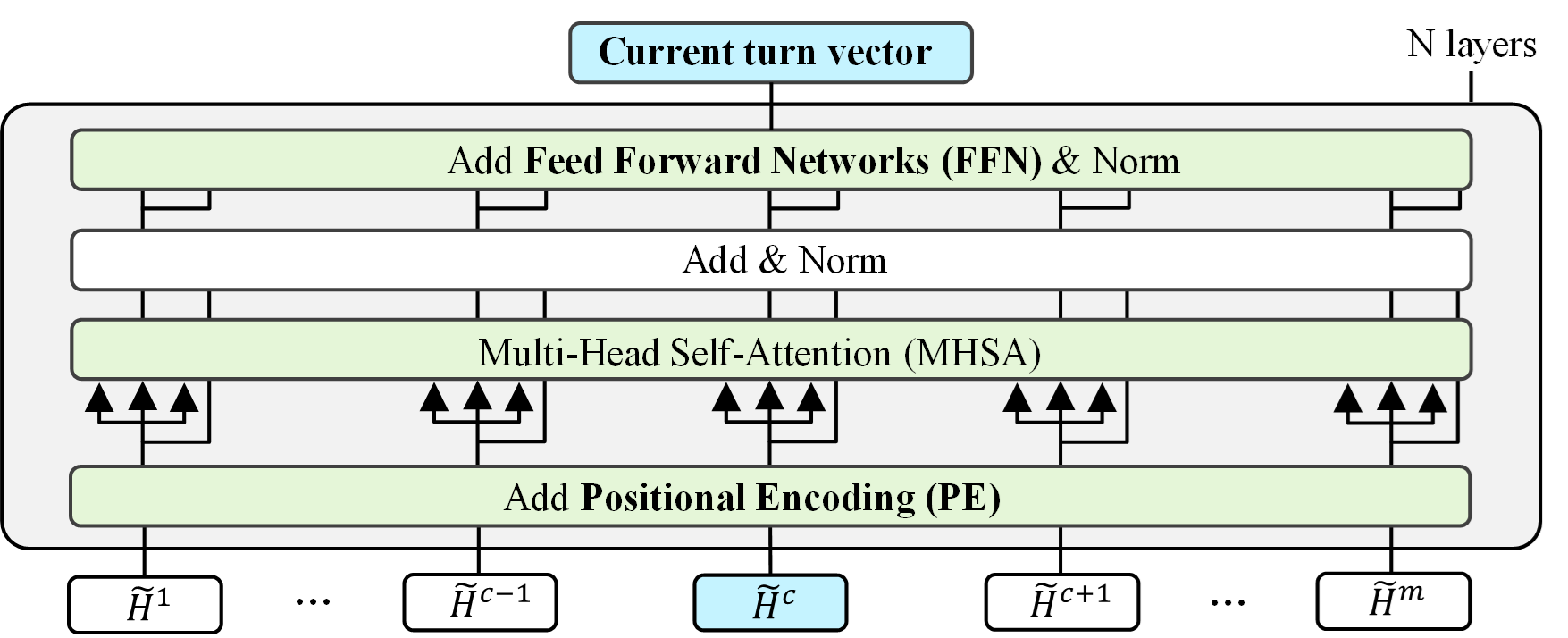}
    \caption{Variants of the TBM structure. A positional encoding (PE) layer of Transformer is placed in front of MHSA and a feed forward network (FFN) is placed in back.}
    \label{fig:fig6}
\end{figure}

\subsection{Comparison with related work using speaker information}
\label{sec:sec5-4}
TBM performs MHSA on turn-based vectors and does not use speaker information, as stated in Section \ref{sec:sec3-2}. Herein, we applied speaker information to TBM in two ways.

The first way involved adding special tokens corresponding to speaker IDs to the input sequence, as is done in TUCORE-GCN. For instance, when the speaker IDs are 0 and 1, the speaker token becomes [SPK0] or [SPK1], respectively. Note that while EmoBERTa uses real names as speaker IDs, we added speaker tokens to use the same IDs as the original data as accurate as possible.

The other way was to attend only to the turns of the target speakers. This is called local attention, and it was originally applied to an encoder-decoder \cite{DBLP:journals/corr/LuongPM15}.

In Table \ref{tab:tab5}, the ``Speaker token`` column shows results for speaker IDs, while the ``Speaker attention`` and ``Listener attention`` columns show results for the attention on turns with the same speaker as the current one and on listener turns. The performance of TBM was significantly higher than that of TBE. However, neither way of adding speaker information showed any advantage. On the contrary, the addition of speaker tokens caused a noticeable loss in performance.

The results also show that it is better to attend to all turns than to limit the attention to certain turns. Moreover, DialogXL reported that global attention contributed more than speaker or listener attention. Accordingly, the results suggests that the high performance of TBM is based on global attention.

\begin{table}[h!]
\footnotesize
\centering
\begin{tblr}{
  width = \linewidth,
  colspec = {Q[2.2]Q[1.1]Q[1.4]Q[1.4]},
  columns = {halign=c, valign=m},
  cell{1}{1} = {r=2, c=1}{halign=c, valign=m},
  cell{1}{2} = {r=2, c=1}{halign=C, valign=m},
  cell{3}{1} = {r=3, c=1}{halign=c, valign=m},
  cell{6}{1} = {r=3, c=1}{halign=c, valign=m},
  cell{9}{1} = {r=3, c=1}{halign=c, valign=m},
  hline{1,3,12} = {-}{},
  hline{6,9} = {2-4}{},
}
{\bf{Transformer}\\\bf{Structure}} & {\bf{\# layers}\\\bf{, heads}} & \bf{IEMOCAP}  & \bf{MELD} \\
                              &                           & \bf{W-Avg F1} & \bf{W-Avg F1} \\
{TBM\\(MHSA)} & 2, 4  & \bf{68.50} & \bf{65.87} \\
                  & 4, 8  & 67.92     & 65.37 \\
                  & 6, 16 & 67.63     & 65.54 \\
\ + Positional Encoding & 2, 4  & 67.97 & 65.86 \\
                  & 4, 8  & 67.64     & 65.27 \\
                  & 6, 16 & 66.93     & 65.50 \\
\ + Feed Forward Network & 2, 4  & 67.64 & 65.49 \\
                  & 4, 8  & 67.47     & 65.54 \\
                  & 6, 16 & 67.66     & 65.74
\end{tblr}
\caption{Comparison of TBM model and TBM with the additional Transformer components. Adding PE and FFN did not give any improvements. This means that MHSA contributed to the high performance of TED.}
\label{tab:tab6}
\end{table}

\subsection{Variants of TBM Structure}
\label{sec:sec5-5}
The MHSA of TBM is the same as that of the vanilla Transformer. Because Transformer-based models use positional encoding (PE) and a feed forward network (FFN), we decided to examine the effect of adding them to TBM in the same way as in the vanilla Transformer as shown in Figure \ref{fig:fig6}.

Table \ref{tab:tab6} shows the effect of adding these Transformer components and the network size depending on the number of layers and heads. The ``TBM`` row contains results for only MHSA, while the ``Positional Encoding`` and ``Feed Forward Networks`` rows show results for TBM with these layers. The ``\# layers`` and ``\# heads`` columns respectively indicate the number of layers of the above components and the number of heads of MHSA. Note that CUST used only the past turns.

The results show that there is no advantage to adding PE and FFN, which are token-based components. Therefore, the results show that TBM works better with turn-based components than with turn-based and token-based components together. Regarding the network parameters, smaller layers and heads are effective. In other words, the complexity of the parameters had no effect on the performance of the turn-based mechanism.

The improvement on IEMOCAP was remarkable, but that on MELD was slight. Regarding this result, more turns would lead to more context and higher performance.
We conclude that, compared with the state-of-the-art methods, TBM, which uses turn-based MHSA between turn-based vectors based on TBE, captures more complicated contexts in the multiple turns.

\section{Conclusion}
\label{sec:sec6}
For modeling multi-turn contexts of conversation, we presented TED, which emphasizes the current turn and explicitly distinguishes each turn by adding the dialogue features into the attention mechanism. The results in four ERC datasets show that TED had good overall performance, while the other models had good performance on certain datasets. Moreover, TED had an advantage in terms of the dialogue features on datasets with a lot of turns, such as IEMOCAP. Further experiments demonstrated the effectiveness of TED’s key components; TBE, TBM, and the dialogue features of the turn position and speaker information. Our priority factor for the distinction of the turns has made it possible to emphasize the emotional target turn and can adapt to diverse datasets by controlling the dialogue features.

\bibliography{custom}
\bibliographystyle{acl_natbib}

\appendix

\section{Attention score (supplementary)}
\label{sec:appendixA}
The attention score \(\alpha_j^{\ t}\) of Formula (\ref{eq:eq5}) is as follows:

\begin{equation}
\begin{split}
    \alpha_j^{\ t} &= Q_k^{\ t} \ \bm{K}_i^{T} \ \sqrt{d^K} \\
    & where \ Q_j^{\ t} = \widetilde{H}^t \ W_j^{\ Q} \ , \ \bm{K}_j ~ \widetilde{\bm{H}}\ W_j^K
    \label{eq:eq11}
\end{split}
\end{equation}

where \({d}^K\) denotes the head dimension to correct the inner product. It is obtained by dividing the model dimension by the number of \({J}\) heads. \({W}_j^{\ Q}\) and \({W}_j^K\) denote learning parameters for the query and key process, respectively. \({Q}_j^{\ t}\) and \({K}_j\) take a linear transformation  to \({d}^K\) from the turn-based vector at the \({t}\)\textsuperscript{th} turn for the query process and the list of the scaled vectors for the key process in all turns, respectively. The superscript \(T\) denotes transpose.

\newpage

\section{Experimental environment}
\label{sec:appendixB}

\begin{table}[h]
\small
\centering
\begin{tblr}{
  width = \linewidth,
  colspec = {Q[1.0]Q[1.0]},
  columns = {halign=c, valign=m},
  hline{1-2, 10} = {-}{},
}
\textbf{Item} & \textbf{Value} \\
OS & Ubuntu18.04 \\
CUDA Version & 10.2 \\
GPU Card & Quadro RTX 8000 \\
GPU RAM Size & 48 GB\\
CPU & Intel® Xeon® Gold 5222, 3.8GHz \\
Deep Learning Framework & PyTorch \\
PyTorch versio & 1.8.1 \\
Python version & 3.7
\end{tblr}
\caption{Computer specifications settings}
\label{tab:tab7}
\end{table}

\begin{table}[h]
\small
\centering
\begin{tblr}{
  width = \linewidth,
  colspec = {Q[1.0]Q[1.0]},
  columns = {halign=c, valign=m},
  hline{1-2, 15} = {-}{},
}
\textbf{HyberParameter} & \textbf{Value} \\
pretrained model & RoBERTa-large \\
optimizer & Adam \\
initial learning rate & 2.00e-06 \\
{decay type of\\learning rate (LR)} & {DailyDiloag: Micro F1\\The others: W-Avg F1} \\
LR decay factor & 0.8 \\
early stopping & not improved within 5 epochs \\
batch size & 4 \\
number of trials & 5 \\
fixed seeds in trials &	1111, 2222, 3333, 4444, 5555 \\
dropout	& 0.1 \\
\# attention layers & 2, 4, 6 \\
\# attention heads & 4, 8, 16 \\
{maximum coefficient\\of normal distribution} & 1.5, 2, 3, 5 \\
\end{tblr}
\caption{Hyperparameter settings}
\label{tab:tab8}
\end{table}

\end{document}